\documentclass[sigconf]{acmart}
\AtBeginDocument{%
  \providecommand\BibTeX{{%
    \normalfont B\kern-0.5em{\scshape i\kern-0.25em b}\kern-0.8em\TeX}}}
\copyrightyear{2020} 
\acmYear{2020} 
\setcopyright{acmcopyright}\acmConference[MM '20]{Proceedings of the 28th ACM International Conference on Multimedia}{October 12--16, 2020}{Seattle, WA, USA}
\acmBooktitle{Proceedings of the 28th ACM International Conference on Multimedia (MM '20), October 12--16, 2020, Seattle, WA, USA}
\acmPrice{15.00}
\acmDOI{10.1145/3394171.3414025}
\acmISBN{978-1-4503-7988-5/20/10}

\usepackage{multirow}
\usepackage{enumitem}
\usepackage{url}
\usepackage{algorithm}
\usepackage{algorithmic}

\begin{document}
\fancyhead{}
\title{One-shot Scene Graph Generation}


\author{Yuyu Guo}
\affiliation{%
	\institution{Center for Future Media, University of Electronic Science and Technology of China, Chengdu, China}
}
\email{yuyuguo1994@gmail.com}
\author{Jingkuan Song}
\authornote{Jingkuan Song is the corresponding author.}
\affiliation{%
	\institution{Center for Future Media, University of Electronic Science and Technology of China, Chengdu, China}
}
\email{jingkuan.song@gmail.com}
\author{Lianli Gao}

\affiliation{%
	\institution{Center for Future Media, University of Electronic Science and Technology of China, Chengdu, China}
}
\email{lianli.gao@uestc.edu.cn}
\author{Heng Tao Shen}
\affiliation{%
	\institution{Center for Future Media, University of Electronic Science and Technology of China, Chengdu, China}
}
\email{shenhengtao@hotmail.com}
\newcommand{\eat}[1]{}
\newcommand{\etal}{\textit{et~al.}}

\begin{abstract}
As a structured representation of the image content, the visual scene graph (visual relationship) acts as a bridge between computer vision and natural language processing. Existing models on the scene graph generation task notoriously require tens or hundreds of labeled samples. By contrast, human beings can learn visual relationships from a few or even one example. Inspired by this, we design a task named One-Shot Scene Graph Generation, where each relationship triplet (e.g., ``dog-has-head'') comes from only one labeled example. The key insight is that rather than learning from scratch, one can utilize rich prior knowledge. In this paper, we propose Multiple Structured Knowledge (Relational Knowledge and Commonsense Knowledge) for the one-shot scene graph generation task. Specifically, the Relational Knowledge represents the prior knowledge of relationships between entities extracted from the visual content, e.g., the visual relationships ``standing in'', ``sitting in'', and ``lying in'' may exist between ``dog'' and ``yard'', while the Commonsense Knowledge encodes ``sense-making'' knowledge like ``dog can guard yard''. By organizing these two kinds of knowledge in a graph structure, Graph Convolution Networks (GCNs) are used to extract knowledge-embedded semantic features of the entities. Besides, instead of extracting isolated visual features from each entity generated by Faster R-CNN, we utilize an Instance Relation Transformer encoder to fully explore their context information. Based on a constructed one-shot dataset, the experimental results show that our method significantly outperforms existing state-of-the-art methods by a large margin. Ablation studies also verify the effectiveness of the Instance Relation Transformer encoder and the Multiple Structured Knowledge.
\end{abstract}




\maketitle

\section{Introduction}
\label{sec.int}
\begin{figure}[t]
	\begin{center}
		\includegraphics[width=1.0\linewidth]{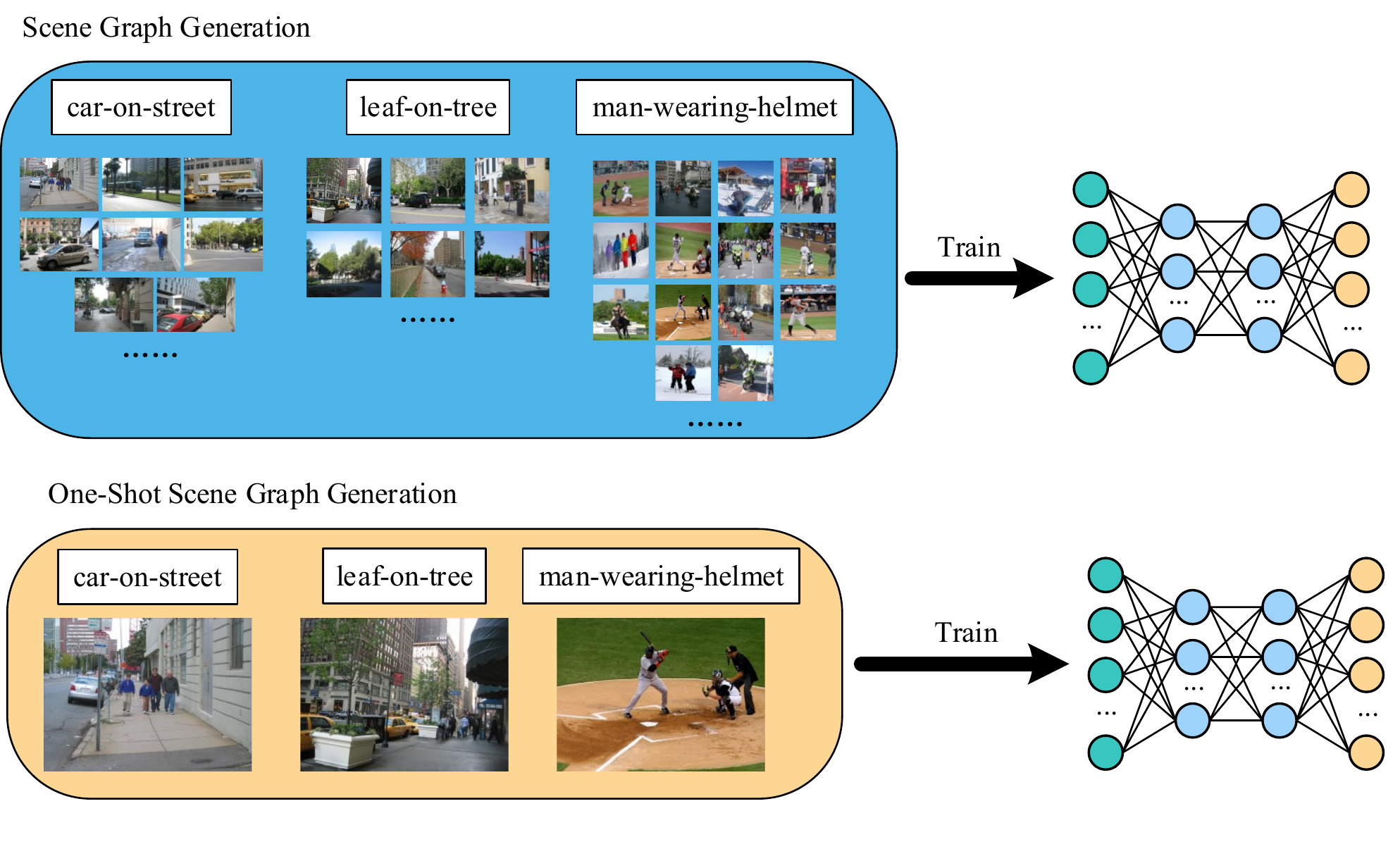}
	\end{center}
	\caption{In this paper, we focus on the one-shot scene graph generation task, where each relationship triplet (e.g., ``dog-has-head'') comes from only one labeled example. For clarity, bounding boxes in the figure are not shown.}
	\label{fig:demo}
\end{figure}
As essential tasks of vision understanding, image classification~\cite{img_class:vgg,img_class:googlenet,img_class:resnet}, image retrieval~\cite{cmm:wjw,imgr:xp1,imgr:shen,imgr:fm}, and object detection~\cite{obj_det:faster,obj_det:fastrcnn,obj_det:yolo} are booming with the development of deep neural networks. However, general attributes of objects, such as category or location, are not adequate to understand image contents. A scene graph, which is a structured representation of the image content, contains not only the semantic and spatial information of objects in images but also relationships between instances. Since the scene graph possesses a wealth of visual contents, the study of scene graph generation facilitates other high-level tasks in the multimedia field~\cite{gan2020foley,gan2020music}, such as visual captioning~\cite{vid_cap:srnn,vid_cap:elr,vid_cap:hieratt} and visual question answering (VQA)~\cite{vqa:xp1,vqa:xp2,vqa:zp1,vqa:zpp2}.

In general, previous methods on scene graph generation focus on the following aspects. 1). How to propose an efficient message-passing mechanism between object features to get the local or global context~\cite{scenegraph:IMP,scenegraph:graphrcnn,scenegraph:motifs,scenegraph:factnet,scenegraph:zoomnet}? 2). How to effectively map visual relationships to a semantic space~\cite{scenegraph:VRD_LP,scenegraph:VTEN,scenegraph:PGAE}? 3). How to design a multi-task network to enhance the scene graph generation task~\cite{scenegraph:denscap,scenegraph:kggan}? Because the semantic space of visual relationships is tremendous, these methods usually require a large number of labeled samples as supervision information.
However, based on rich prior knowledge in the brain, humans can easily overcome this difficulty and learn visual relations from few or even one example.
\begin{figure}[t!]
	\begin{center}
		\includegraphics[width=1.0\linewidth]{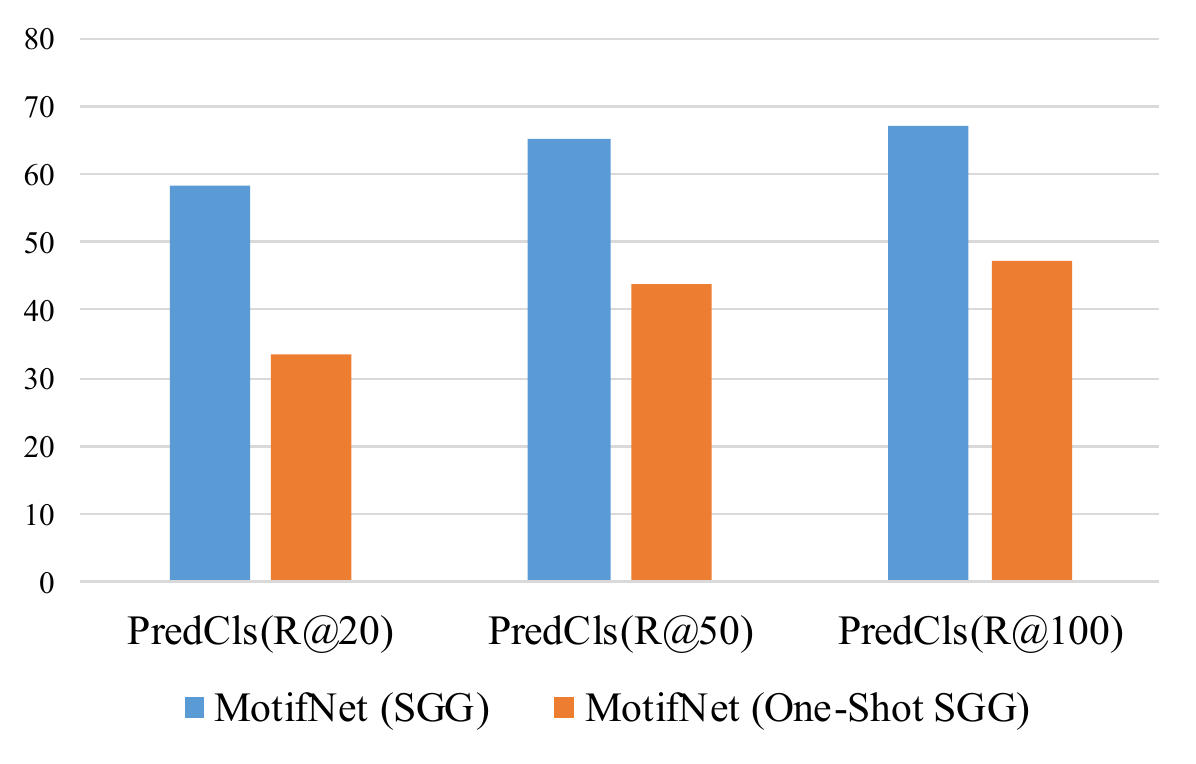}
	\end{center}
	\caption{MotifNet~\cite{scenegraph:motifs} is applied to the  one-shot scene graph generation task. The performances are evaluated on the PredCls (Recall@K) setting. The blue bars denote the performances of MotifNet on the scene graph generation task. The orange bars denote the performances of MotifNet on the one-shot scene graph generation task.}
	\label{fig:motif_drop}
\end{figure}

In order to equip models with the ability to learn visual relationships from one example, we design a new task called One-Shot Scene Graph Generation, where each relationship triplet contains only one annotated example, as shown in Figure~\ref{fig:demo}. Due to a lack of sufficient supervision information, this task is difficult for the previous work, as illustrated in Figure~\ref{fig:motif_drop}. 
Directly applying existing scene graph generation models on this task leads to a significant performance drop. 

As mentioned above, rich prior knowledge helps humans to learn visual relationships from few examples. This suggests that models should pay attention to not only visual information but also other information from prior knowledge~\cite{kg:actr,kg:conceptnet,kg:kgnet}. In this paper, the Multiple Structured Knowledge (Relational Knowledge and Commonsense Knowledge) is introduced into the one-shot scene graph generation task. The Relational Knowledge represents prior knowledge of relationships between entities. For example, there is a high probability that the relationship ``play'' exists between ``person'' and ``dog'', even if the image is not visible. The Commonsense Knowledge precisely locates entities in the commonsense and helps models to reason effectively. When we have known the facts ``horse is an animal'' and ``man can raise horses'', it is natural to infer ``man can raise animals'', even if we have not seen other animals.

In order to handle the one-shot scene graph generation task, the Multiple Structured Knowledge is introduced into our method as following steps: 1) Encoding instance features with an Instance Relation Transformer; 2) Extracting the Relational Knowledge and the Commonsense Knowledge from knowledge bases; 3) Organizing the prior knowledge into graph structures; 4) Encoding the graph-structured knowledge with GCNs; and 5) Combining the GCNs and the Instance Relation Transformer for relationship predicate classification. 
Specifically, motivated by the Transformer~\cite{networks:transformer} structure, we propose the Instance Relation Transformer encoder to capture the relational context among instances in an image. Then the Relational Knowledge is extracted from a relation knowledge base (Visual Genome~\cite{scenegraph:visual_genome}), and the Commonsense Knowledge is obtained from a commonsense knowledge base (ConceptNet~\cite{kg:conceptnet}). These large knowledge bases consist of many loose triplets, and it is unwieldy to obtain knowledge features from these triplets. In this paper, the triplets in the knowledge bases are organized into graph structures. Graph Convolutional Networks (GCNs)~\cite{networks:gcn} encode the graph structures to extract knowledge features. Finally, the outputs of the Instance Relation Transformer and the GCNs are combined to predict relationships between instances. 

The contributions of this paper can be summarized as:
1) To imitate the way human beings understand visual relationships, this work first defines the one-shot scene graph generation task, where the supervision information of each relationship triplet only comes from one labeled example;
2) Relational Knowledge and the Commonsense Knowledge are introduced into the one-shot scene graph generation task. The Relational Knowledge provides the prior knowledge about the relationships of entities, and the Commonsense Knowledge encodes ``sense-making'' knowledge. 
An Instance Relation Transformer encoder is utilized to explore the context information of visual entity for scene graph generation; 
and 3) We collect a new dataset for the one-shot scene graph generation task, where each relationship (subject, predicate, object) contains only one annotated example. Experiments show that our method significantly outperforms existing state-of-the-art methods by a large margin.

\section{Related Work}

\subsection{Scene Graph Generation}
The scene graph defined by Johnson~\etal~\cite{scenegraph:imgrev} is composed of a series of nodes and edges. The nodes are represented by entities in images, which contain categories and locations of entities. The edges are represented by visual relationship predicates between entities, such as ``on'', ``in'' and ``under''. 
As mentioned above, the previous methods for the scene graph generation task are roughly based on the following three perspectives: 1). Building a semantic space for visual relationships from a language model~\cite{scenegraph:VRD_LP} or a Translation Embedding~\cite{scenegraph:VTEN}; 2). Extracting contextual information from bipartite sub-graphs~\cite{scenegraph:motifs}, global architectures~\cite{scenegraph:motifs}, or tree structures~\cite{scenegraph:treelstm}; 3). Combining with multi-tasks, such as dense captioning~\cite{scenegraph:denscap} and image reconstruction~\cite{scenegraph:kggan}.
Most of these efforts need a large-scale annotated dataset, which requires much labor. However, human beings can understand visual relationships from few examples. To imitate such ability, we design a one-shot scene graph task and introduce rich knowledge from human beings to handle this task.
\begin{figure*}[hbt!]
	\begin{center}
		\includegraphics[width=0.85\linewidth]{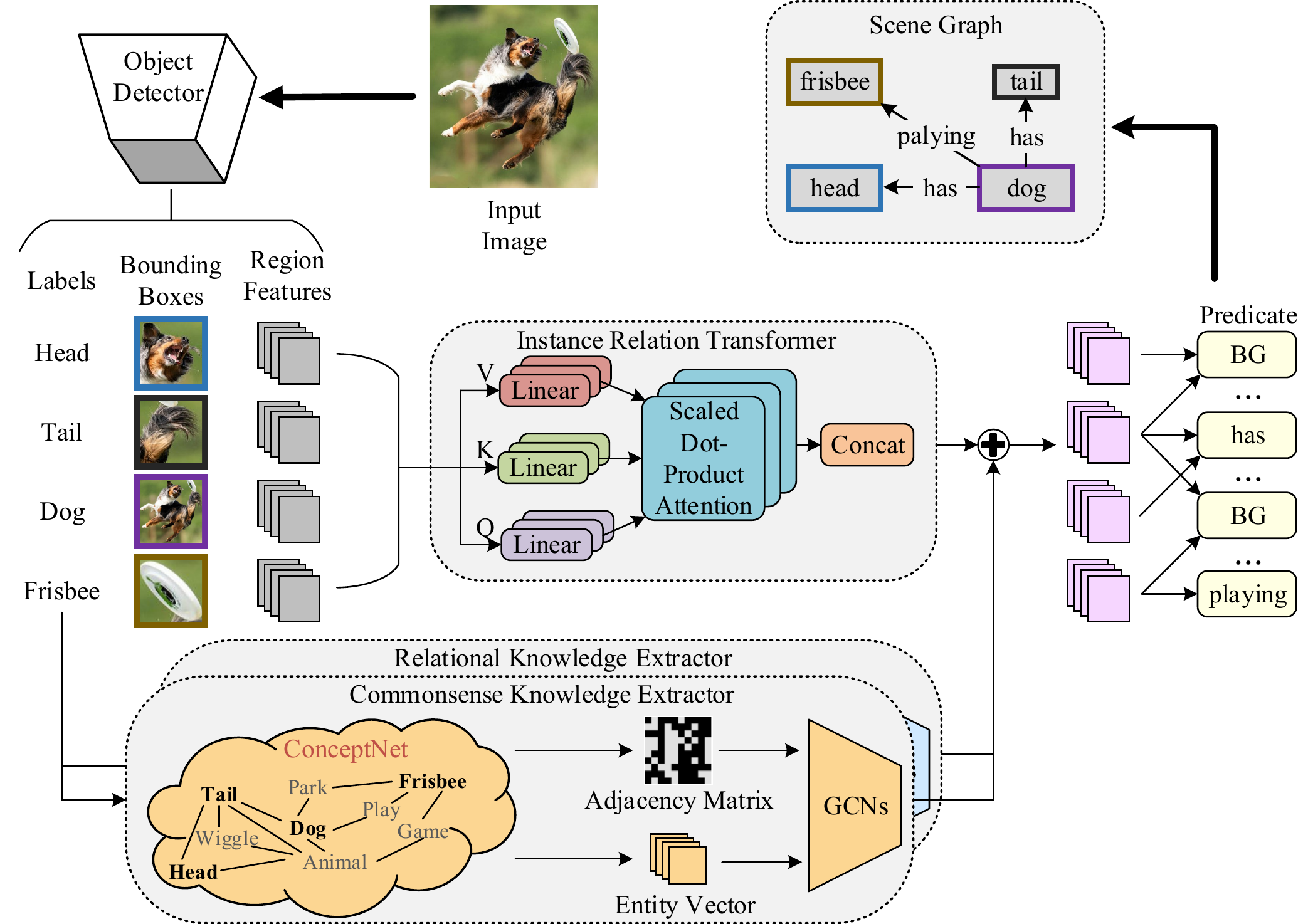}
	\end{center}
	\caption{The framework of our method. Given an image, we first detect the instances in the image with Faster R-CNN. Then the Instance Relation Transformer network is proposed to explore the contextual information among instances. Next, the Relational Knowledge Extractor extracts the relational knowledge features from Visual Genome, and the Commonsense Knowledge Extractor extracts the commonsense knowledge features from ConceptNet. Finally, the relationship predicates are predicted with outputs of the Instance Relation Transformer, the Relational Knowledge Extractor, and the Commonsense Knowledge Extractor.}
	\label{fig:framework}
\end{figure*}

\subsection{One-Shot learning}
Many works~\cite{oneshot:feifei,oneshot:hierabay,oneshot:mulatt} have shown that the machine can understand a wealth of information from one example. Feifei~\etal~\cite{oneshot:feifei} explored general knowledge from learned categories and utilized a Bayesian method to implement one-shot learning of object categories. However, in their work, the prior knowledge contains only three categories. This weakens the generalization of the method. To alleviate this issue, Salakhutdinov~\etal~\cite{oneshot:hierabay} collected a set of super-categories to represent different priors for new categories and constructed a hierarchical Bayesian model for learning from one example. Both of these Bayesian methods lack powerful image features. Due to the success of deep learning in image representation, Koch~\etal~\cite{oneshot:siann} proposed a siamese neural network that combines the convolutional networks and the metric learning strategy. Orthogonal to the above methods, Wang~\etal~\cite{oneshot:mulatt} designed a multi-attention network that generates the image features from the category semantic embedding. Different from the above works, our work focuses on the scene graph generation task under the one-shot environment and adopts the rich knowledge from knowledge bases to support this task.

\section{Approach}
In this paper, we introduce the Multiple Structured Knowledge into the one-shot scene graph generation task. The framework of our method is depicted in Figure~\ref{fig:framework}. The proposed method contains four main components: 1) An object detector; 2) An Instance Relation Transformer encoder; 3) A Relational Knowledge extractor; and 4) A Commonsense Knowledge extractor. In this section, we first define the one-shot scene graph generation task and then introduce each part from inputs to outputs. 

\subsection{Problem Definition of One-Shot Scene Graph Generation}
Given an image, a scene graph is defined as a set of nodes and edges, where the nodes represent instances in the image, while the edges represent the relationships between instances. The scene graph can be divided into relationship triplets <subject, predicate, object>, where the subject and object are the instances detected by an object detector. This task requires a model to predict relationship predicates between instances. 

For the one-shot scene graph generation task, the ground truth of each relationship triplet <subject, predicate, object> contains only one labeled example as supervision information. Specifically, a one-shot dataset from the Visual Genome dataset is built to support this task. 
During the construction process, we first initial the one-shot dataset $D$ with none. Next, all images in the Visual Genome dataset are checked.
An image faces two situations: 
1). When an image contains relationship triplets that do not appear in dataset $D$ yet, we add the image and the corresponding annotations of the relationship triplets to dataset $D$.
2). When all relationship triplets of an image have appeared in $D$, the dataset $D$ skips the image. The one-shot dataset $D$ is collected to verify our method on the one-shot scene graph generation task.

\subsection{Object Detector}
\label{subsec:obj_det}
In this paper, we adopt Faster R-CNN to generate $n$ instances, which include the following information: 
\begin{itemize}
	\item Label probabilities $L = \{l_1,...,l_n\}$, where $L \in \mathbb{R}^{n \times d^a}$ and $d^a$ is the  number of instance categories;
	\item Bounding boxes $B = \{b_1,...,b_n\}$, where $B \in \mathbb{R}^{n \times 4}$;
	\item Object features $F = \{f_1,...,f_n\}$, where $F \in \mathbb{R}^{n \times d^c}$ and $d^c$ is the feature dimension;
	\item Union object features $U = \{u_{1,1},...,u_{n,n}\}$, where $U \in \mathbb{R}^{n \times n \times d^c}$. 
\end{itemize}
The object feature $f_i$ is extracted from bounding box $b_i$. The label probability $l_i$ is generated with Faster R-CNN from bounding box $b_i$. The union object feature $u_{i,j}$ is extracted from the union bounding box of $b_i$ and $b_j$.
\subsection{Instance Relation Transformer}
Generating a complete scene graph requires not only the visual features of instances but also the contextual information. For example, when we know ``people feed horses'', we should also increase the confidence of ``horses on the ground''. However, the isolated features $F$ from Faster R-CNN ignore the surrounding context, and it is necessary to use an effective strategy to get the context in an image for the scene graph generation task. Most of the previous methods~\cite{scenegraph:motifs,scenegraph:treelstm,scenegraph:IMP} utilize RNNs to obtain the global or local context. Nevertheless, as stated by Vaswani~\etal~\cite{networks:transformer}, RNNs have defects in parallelization, computational complexity, and long-term dependence. In this paper, in order to understand the context of relationships effectively, the Transformer network is utilized for encoding the instance features. Since the Transformer structure can explore the relationship among inputs, this structure is suitable for capturing the relational context in an image. 

We construct the Instance Relation Transformer and generate the relation context features $M \in \mathbb{R}^{n \times d^z}$ by applying the Transformer structure to the instance information, i.e., visual features, label embeddings, and position embeddings:
\begin{equation}
\begin{array}{l}
M = {\rm{Transformer}}([F,E^g,E^v];W^z)~,
\end{array}
\end{equation}

where $W^z$ is a parameter set in the Transformer. $E^g$ and $E^v$ are the embedding vectors from label probabilities $L$ and bounding boxes $B$, respectively. $F$ is the instance feature mentioned in Section~\ref{subsec:obj_det}. $[:,:,]$ means the concatenate operation. 
\subsection{Relational Knowledge Extractor}
\label{sec.spk}
As discussed in the Section~\ref{sec.int}, the Multiple Structured Knowledge (Relational Knowledge and Commonsense Knowledge) is introduced into the one-shot scene graph generation task. The Relational Knowledge extractor and the Commonsense Knowledge extractor are designed to capture the relational knowledge features and commonsense knowledge features, respectively.

We first introduce the Relational Knowledge, which contains the prior knowledge of the relationship between entities in the visual space. Specifically, the Relational Knowledge is extracted from a relation dataset: Visual Genome. Visual Genome bridges the gap between computer vision and natural language processing, and can be used for many tasks, such as VQA, image captions, and scene graphs. In particular, we use scene graph labels filtered by Xu~\etal~\cite{scenegraph:IMP} as our knowledge base. A series of triplets <subject, predicate, object>, e.g., <pillow, on, bed>, represent the scene graph in Visual Genome. All triplets in the training dataset are organized into a Relational Knowledge base $K^v$, which contains $200$ entity categories $C^{q}$ that include subjects, objects, and predicates. The structured knowledge is represented as a set of adjacency matrices and entity vectors. In this work, two boolean adjacency matrices $A^o \in \mathbb{R}^{200 \times 200}$ and $A^p \in \mathbb{R}^{200 \times 200}$ are constructed to represent the Relational Knowledge. The boolean adjacency matrix $A^o$ represents whether there are triplets between entity categories. To capture predicate information, the adjacency matrix $A^p$ focuses on the relationship between objects/subjects and predicates. For example, the relationship triplet <pillow, on, bed> is contained in the Relational Knowledge base $K^v$, and $x$, $y$ and $z$ are the indexes of ``pillow'', ``bed'', and ``on'' in $C^{q}$, respectively. Then the element $a^o_{x,y}$ of the boolean adjacency matrix $A^o$ is set to $1$, and the elements $a^p_{x,z}$ and $a^p_{z,y}$ of the boolean adjacency matrix $A^p$ are both set to $1$.

After obtaining the boolean adjacency matrices, the Word2Vector method is adopted to extracts the entity vectors $E^p$ according to the corresponding vocabularies in the categories $C^{q}$. In order to capture the structured information in the adjacency matrices $A^o$ and $A^p$, Graph Convolutional Networks (GCNs)~\cite{networks:gcn} are utilized for encoding the entity vectors $E^p$:
\begin{equation}
\begin{array}{llll}
{O^{v1}} &= &{\rm{GCNs}}({E^p},{A^o};{W^{g1}})~,\\
{O^v} &= &{\rm{GCNs}}({O^{v1}},{A^p};{W^{g2}})~,
\end{array}
\label{equ.vggcn}
\end{equation}
where $W^{g1}$ and $W^{g2}$ are parameters in GCNs. We get the relational knowledge features $O^v \in \mathbb{R}^{200 \times d^z} $ of the category set $C^{q}$. Finally, we find the indexes of label $L$ (mentioned in Section~\ref{subsec:obj_det}) in $C^{q}$, and then use these indexes to select the corresponding features from $O^v$ to form the Relation Knowledge features $P^v \in \mathbb{R}^{n \times d^z}$ of label $L$.

\subsection{Commonsense Knowledge Extractor}
The Commonsense Knowledge defines the exact meaning of entities, which can assist the cognition and reasoning of the model. Inspired by~\cite{kg:conceptnet}, a commonsense knowledge base (ConceptNet) is used to obtain the Commonsense Knowledge.
ConceptNet contains many loose triplets <head, relation, tail>, such as <dog, desires, play> and <frisbee, usedfor, play>. The head and the tail represent a head concept and a tail concept in ConceptNet, respectively. The relation represents a semantic relationship, such as ``desires'', ``has property'' and ``is used for''.
The commonsense information in ConceptNet facilitates a model to understand the definitions of objects. However, the ConceptNet dataset is large and hard to be used directly. It is necessary to refine and filter the ConceptNet dataset. In order to increase the density of ConceptNet, the original relation categories are deleted and merged following the approach from Lin~\etal~\cite{kg:kgnet}.

As mentioned above, it is difficult to mine valuable information from loose triplets in ConceptNet. Previous work~\cite{scenegraph:kggan} utilizing LSTM to directly encode the loose triplets can not effectively extract the structured knowledge among triplets. In our work, we build a subgraph from simple paths constructed by the loose triplets and use GCNs on the subgraph to extract the knowledge features. 
To construct the subgraph, the method~\cite{kg:kgnet} is adopted in ConceptNet to find and prune simple paths constructed by triplets. We first locate the instance labels of $L$ (mentioned in Section~\ref{subsec:obj_det}) in the concepts of ConceptNet. Then, for each label pair of $i$-th label $l_i$ and  $j$-th label $l_j$, all simple paths between $l_i$ and $l_j$ along triplets in ConceptNet are checked. If the length of a simple path is shorter than five, it is reserved. Otherwise, it is discarded. For path pruning, each triplet in a path is rated with the TranSE~\cite{kg:transe} method first. The score of each path is the product of the scores of triplets in the path. Then the paths with scores less than 0.15 are filtered out. Finally, for an image, the concepts in the filtered paths of all label pairs are organized into a new category set $C^c$. A boolean adjacency matrix $A^c$ is used to indicate whether two concepts are adjacent in the filtered paths.

Similar to Relational Knowledge Extractor, the Word2Vector method extracts entity vectors $E^c$ from the category set $C^c$, and GCNs capture commonsense knowledge features from $A^c$ and $E^c$:	
\begin{equation}
\begin{array}{llll}
{O^c} &= &{\rm{GCNs}}({E^c},{A^c};{W^{g3}})~.\\
\end{array} 
\label{equ.cptgcn}
\end{equation}
$O^c \in \mathbb{R}^{|C^c| \times d^z} $ is the commonsense knowledge feature of $C^c$, and $|C^c|$ is the number of elements in the category set $C^c$. The commonsense knowledge features $P^c \in \mathbb{R}^{n \times d^z}$ of the instance labels are extracted from $O^c$ according to the indexes of $L$ in $C^c$.

Until now, we can obtain the semantic features of the Multiple Structured Knowledge ($P^v$ and $P^c$).
\subsection{Predicate Classification}
In order to represent the detected instance, we sum the outputs from the Instance Relation Transformer ($M$), the Relational Knowledge extractor ($P^v$), and the Commonsense Knowledge extractor ($P^c$):
\begin{equation}
\begin{array}{llll}
E^r = P^v + P^c + M~.\\
\end{array}
\label{equ.add^thr}
\end{equation}
Because the same instance is inconsistent in the subject space and the object space, $E^r$ is mapped to the subject space and object space with fully connected (FC) layers:
\begin{equation}
\begin{array}{llll}
E^s & = & {\rm{FC}}(E^r;W^s)~, \\
E^o & = & {\rm{FC}}(E^r;W^o)~,
\end{array} 
\label{equ.so_fc}
\end{equation}
where $W^s$ and $W^o$ are parameters. $E^s=\{e^s_1,e^s_2,...,e^s_n\}$ and $E^o=\{e^o_1,e^o_2,...,e^o_n\}$.

The DisMult~\cite{kg:dist_mult} method predicts relation predicate between $i$-th instance and $j$-th instance:
\begin{equation}
\begin{array}{ll}
r_{i, j, k} = (e^s_i \circ u_{i,j})^T W^r_k (e^o_j \circ u_{i,j}) + b^r_{i,j, k}~,
\end{array}
\label{equ.rel_dismult} 
\end{equation}
where $r_{i, j, k}$ is the probability that the $k$-th relation predicate exists between the $i$-th instance and the $j$-th instance. $W^r_k$ is a diagonal parameter matrix for the $k$-th relation predicate.  $b^r_{i,j,k}$ is a frequency baseline proposed by~\cite{scenegraph:motifs}. In addition, $E^s$ and $E^o$ are also used to predict the probability $r'_{i, j}$ that indicates the probability of non-background relationship between the $i$-th instance and the $j$-th instance:
\begin{equation}
\begin{array}{ll}
r'_{i, j} = (e^s_i \circ u_{i,j})^T W^{r'} (e^o_j \circ u_{i,j}) + b^{r'}_{i,j}~.
\end{array}
\label{equ.rel_dismult1} 
\end{equation}
If $r'_{i, j} = 1$, there is a non-background relationship between the $i$-th instance and the $j$-th instance.
Finally, we apply the softmax function to $r_{i, j, k}$ and  $r'_{i, j}$, and use the cross entropy function to learn the parameters of the model.

\section{Experiments}
In this section, we conduct experiments on two datasets: the Visual Genome dataset and the One-Shot Visual Genome dataset. Firstly, in order to verify whether a model can learn each visual relationship from one example, our method and some existing methods are evaluated on the one-shot scene graph task. Secondly, a detailed ablation study is conducted on the one-shot scene graph generation task to verify the effectiveness of each component. Then, we show that our method can also handle the scene graph generation task properly. Finally, some visual results are shown for qualitative analysis. 

\subsection{Datasets}
\textbf{Visual Genome.} For modeling relationships in our visual world, Krishna~\etal~\cite{scenegraph:visual_genome} collected a dense annotation dataset called Visual Genome, where each image is annotated with objects, attributes, and relationships. The Visual Genome dataset contains over $100K$ images. Each image contains about $21$ instances, $18$ attributes, and $18$ relationship triplets. However, it is difficult for models to learn stable information due to a lot of low-quality annotations in this dataset. Therefore, Xu~\etal~\cite{scenegraph:IMP} brought forward an approach to filter the low-quality annotations, which is widely used in other works~\cite{scenegraph:cmat,scenegraph:graphrcnn,scenegraph:treelstm,scenegraph:motifs}.

Each image contains about $12$ instances and $6$ relationship triplets in the filtered Visual Genome dataset. The dataset contains $150$ instance categories and $50$ predicate categories in total. The training set and testing set account for $70\%$ and $30\%$ of the entire dataset, respectively. Moreover, following previous works~\cite{scenegraph:motifs,scenegraph:kggan}, the validation set consists of 5k images from the training set. The previous works~\cite{scenegraph:denscap,scenegraph:kggan,scenegraph:zoomnet} also proposed other different filtering strategies. 
This paper ignores the different filtering strategies and employs the filtering strategy proposed by Xu~\etal~\cite{scenegraph:IMP} to verify our method.

\textbf{One-Shot Visual Genome.} 
To handle the one-shot scene graph generation task, we collect an extreme training dataset from the Visual Genome dataset, where each relationship triplet only appears once. A relationship triplet set is constructed from the training set proposed by~\cite{scenegraph:IMP}. The triplet set contains about $29K$ relationship triplets in total. For each triplet, we randomly select an image for forming the One-Shot Visual Genome dataset. Finally, the dataset contains about $18K$ images. Since an image may contain several different triplets, the number of images is less than the number of triplets. It is worth emphasizing that we only use this dataset to train models, while the test dataset remain the same as the test set of Visual Genome.
\begin{table*}[t]
	\centering
	\caption{Existing methods decline significantly on the one-shot scene graph generation task. SGG denotes scene graph generation. OSSGG denotes one-shot scene graph generation. MSK denotes the Multiple Structured Knowledge. IRT denotes the Instance Relation Transformer.}
	\scalebox{0.9}{	
	\begin{tabular}{c|c|c|c|c|c|c|c|c|c|c}
		\hline
		\multirow{2}{*}{Dataset}
		& \multirow{2}{*}{Method} & \multicolumn{3}{c|}{PredCls}
		& \multicolumn{3}{c|}{SGCls} & \multicolumn{3}{c}{SGDet}  \\ \cline{3-11} 
		& & R@20    & R@50   & R@100   & R@20    & R@50   & R@100   & R@20   & R@50   & R@100   \\ \hline
		\multirow{4}{*}{SGG}
		& FREQ+OVERLAP~\cite{scenegraph:motifs}       
		& 53.6          & 60.6          & 62.2                 
		& 29.3          & 32.3          & 32.9        
		& 20.1          & 26.2          & 30.1          \\ \cline{2-11}
		
		& IMP+~\cite{scenegraph:IMP}     
		& 52.7          & 59.3          & 61.3                   
		& 31.7          & 34.6          & 35.4           
		& 14.6          & 20.7          & 24.5  \\\cline{2-11}
		& MotifNet~\cite{scenegraph:motifs}                                                               
		& 58.5    & 65.2    & 67.1
		& 32.9    & 35.8   & 36.5  
		& 21.4    & 27.2   & 30.3      \\ \cline{2-11}  
		& IRT (Ours)                                   & 60.3   & 66.8   & 68.5 
		& 33.9   & 36.9   & 37.5   
		& 21.9   & 27.8   & 31.0     \\ \hline \hline
		\multirow{5}{*}{OSSGG}
		& FREQ+OVERLAP\footnotemark[\getrefnumber{foot1}]  
		& 5.0          & 8.8         & 11.3                 
		& 3.3          & 5.1          & 5.9        
		& 1.4          & 2.8          & 4.0          \\ \cline{2-11}
		& IMP+\footnotemark[\getrefnumber{foot1}]  
		& 36.4          & 45.3          & 48.3                   
		& 19.7          & 23.8          & 25.0           
		& 4.5          & 8.6          & 12.6  \\\cline{2-11}
		& MotifNet\footnotemark[\getrefnumber{foot1}] 
		& 33.5    & 43.6   & 47.1                 
		& 17.4    & 22.0   & 23.4
		& 6.2     & 9.4    & 11.8         \\ \cline{2-11}
		& IRT (Ours)
		& 37.7   & 46.1   & 48.9   
		& 21.2   & 25.2   & 26.4    
		& 6.8    & 11.1   & 15.0     \\ \cline{2-11}
		& IRT+MSK (Ours)
		& \textbf{41.3}   & \textbf{49.5}   & \textbf{52.2}                         
		& \textbf{23.0}   & \textbf{26.9}   & \textbf{28.1} 
		& \textbf{7.1}    & \textbf{11.5}   & \textbf{15.5}     \\ \hline 
	\end{tabular}}
	\label{tab.res_savg}

\end{table*}

\subsection{Implementation Details}

For the object detector, Faster R-CNN with RoI Align provided by Zellers~\cite{scenegraph:motifs} is used to detect instances in images, and its parameters are frozen. Besides the Instance Relation Transformer mentioned above for predicting relationship predicates, we utilize another Instance Relation Transformer to refine the instance labels on the scene graph generation task. The depth and width of the Instance Relation Transformers are both set to $6$, and the dimension is set to $768$. For experiments on the one-shot scene graph generation task, the depth and width of the Instance Relation Transformer are both set to $12$ and the dimensions to $768$. The numbers of layers of GCNs in Equation~\ref{equ.vggcn} and Equation~\ref{equ.cptgcn} are 2, and the dimensions of GCNs are $768$. 

We use the SGD method to learn the parameters. The learning rate and the batch size are set as $5\times10^{-3}$ and $16$, respectively. The maximum number of epochs is 50. 
Python and Pytorch are adopted to build our model. All the experiments are carried out on the Ubuntu system with 256 GB RAM, a Titan Xp (12 GB) GPU, and 4 Intel(R) Xeon(R) E5-2650 CPUs.

\subsection{Evaluation Strategies}
Following previous works, we use three setups (PredCls, SGCls and SGDet) to evaluate our method. The PredCls (predicate classification) task predicts relationship predicates with the ground truths of bounding boxes and categories in the test phase. The SGCls (scene graph classification) task allows models to employ the ground truths of bounding boxes in the test phase. The SGDet (scene graph detection) setup requires the model to generate bounding boxes, categories, and relationship predicates in the test phase without any ground truths. All three cases are evaluated by Recall@K (R@K, K=20,50,100). In this paper, we show all the results with graph constraint, i.e., each instance pair produces a relationship triplet.

\subsection{Experimental Results on the One-Shot Scene Graph Generation task}

As mentioned above, human beings can learn stable information from few samples or even one sample. In order to check whether models possess such ability, we show the experimental results of our method and existing methods on the one-shot scene graph generation (OSSGG) task.

Due to the lack of sufficient supervision information, the OSSGG task is more difficult than the scene graph generation task. Directly applying existing methods to the OSSGG task can lead to a significant performance drop, as shown in Table~\ref{tab.res_savg}.
It can be seen that the decline of FREQ+OVERLAP is the most severe since it just relies on the bias of the dataset. It is noteworthy that the decline rate of our IRT is lower than that of MotifNet. For example, on $R@20$ of PredCls, the rate of decline for our IRT is $37.5\%$ ($(60.3-37.7)/60.3$), and the rate of decline for MotifNet is $42.7\%$ ($(58.5-33.5)/58.5$). This shows that the contextual information extracted by our IRT is more robust than MotifNet, which uses Bi-LSTM to extract the global context. 

Moreover, the Multiple Structured Knowledge enhances the Instance Relation Transformer on the OSSGG task. Under the conditions of PredCls and SGCls, the Multiple Structured Knowledge notches up high growth rates ($(41.3-37.7)/37.7=9.5\%$ and $(23.0-21.2)/21.2=8.5\%$). On the SGDet task, the growth rates brought by the Multiple Structured Knowledge are marginal, because the SGDet setting depends heavily on object detection. However, the emphasis of our work is predicate prediction rather than object detection. These experiments verify the validity of the Multiple Structured Knowledge on the OSSGG task.

In general, compared with the above methods, our method achieves the best on the scene graph generation task as well as the one-shot scene graph generation task.

\subsection{Ablation Study}
In order to profoundly analyze our method and illustrate the effectiveness of each component, we conduct a set of detailed ablation experiments on the one-shot scene graph generation task.

\textbf{The role of Multiple Structured Knowledge.}
To further illustrate the role of Multiple Structured Knowledge, we test the effects of the Multiple Structured Knowledge and visual features, as shown in Table~\ref{tab.res_visorknow}. Using ResNet enhances the results of IRT with an increase of $2.8\%$ on R@20 compared with VGG. This means that a powerful visual feature can improve the robustness of IRT on the OSSGG task. Moreover, the results of IRT(V)+MSK are better than IRT(R) ($41.3$ vs. $40.5$). These results illustrate the effectiveness of our Multiple Structured Knowledge. When we use both ResNet and Multiple Structured Knowledge, the model achieves the best. This shows the adaptability of our Multiple Structured Knowledge.

\begin{table}[t]
	\centering
	\caption{Ablation study of visual features and Multiple Structured Knowledge. V denotes VGG-16. R denotes ResNet-101.}
	\scalebox{0.9}{	
	\begin{tabular}{c|c|c|c}
		\hline
		\multirow{2}{*}{\begin{tabular}[c]{@{}c@{}}Method\end{tabular}} & \multicolumn{3}{c}{PredCls} \\ \cline{2-4} & R@20  & R@50   & R@100   \\ \hline
		IRT(V)                                                & 37.7   & 46.1   & 48.9    \\ \hline
		IRT(R)                                         
		& 40.5   & 48.6   & 51.2    \\ \hline
		IRT(V)+MSK                        
		& 41.3   & 49.5   & 52.2 \\ \hline 
		IRT(R)+MSK                       
		& \textbf{41.8}   & \textbf{51.3}   & \textbf{54.3}  \\ \hline 
	\end{tabular}}
	\label{tab.res_visorknow}
\end{table}

\begin{table}[t]
	\centering
	\caption{Ablation study of the Commonsense Knowledge and the Relational Knowledge. CK denotes the Commonsense Knowledge from the ConceptNet dataset. RK denotes the Relational Knowledge from the Visual Genome dataset. }
	\scalebox{0.9}{	
	\begin{tabular}{c|c|c|c|c|c}
		\hline
		\multirow{2}{*}{CK} & \multirow{2}{*}{RK} & \multirow{2}{*}{IRT} & \multicolumn{3}{c}{PredCls} \\ \cline{4-6} 
		&                         &                              & R@20    & R@50    & R@100    \\ \hline
				&                         & $\surd$                            & 37.7    & 46.1    & 48.9     \\ \hline

		& $\surd$                       & $\surd$                            & 39.9    & 48.2    & 51.0     \\ \hline
		$\surd$                            & $\surd$                       & $\surd$                            & \textbf{41.3}    & \textbf{49.5}    & \textbf{52.2}     \\ \hline
	\end{tabular}}
	\label{tab.ablation_spk}
\end{table}

\begin{table}[t]
	\centering
	\caption{Ablation study of the structured knowledge from Visual Genome. $A^o$ is the adjacency matrix with instance categories mentioned in Section~\ref{sec.spk}. $A^p$ is the adjacency matrix with predicate categories.}
	\scalebox{0.9}{	
	\begin{tabular}{c|c|c|c|c|c}
		\hline
		\multirow{2}{*}{$A^p$} & \multirow{2}{*}{$A^o$} & \multirow{2}{*}{IRT} & \multicolumn{3}{c}{PredCls} \\ \cline{4-6} 
		&                         &                              & R@20    & R@50    & R@100    \\ \hline
	    &   & $\surd$   & 37.7    & 46.1    & 48.9     \\ \hline

        &  $\surd$ & $\surd$   & 38.7    & 47.5    & 50.3     \\ \hline
      	 \eat{$\surd$	&  & $\surd$   & 39.4    & 48.0    & 50.7     \\ \hline}
$\surd$     & $\surd$  & $\surd$ & \textbf{39.9}    & \textbf{48.2}    & \textbf{51.0}    \\ \hline
	\end{tabular}}
	\label{tab.ablation_spkvg}
\end{table}
\eat{
\begin{table}[t]
	\centering
	\caption{Ablation studies of the structured knowledge from Visual Genome. $A^o$ is the adjacency matrix without predicate categories mentioned in Section~\ref{sec.spk}. $A^p$ is the adjacency matrix with predicate categories mentioned in Section~\ref{sec.spk}. IRT denotes Instance Relation Transformer.}
	\begin{tabular}{c|c|c|c}
		\hline
		\multirow{2}{*}{Methods}      & \multicolumn{3}{c}{PredCls} \\ \cline{2-4} 
		& R@20    & R@50    & R@100    \\ \hline
		IRT                   & 37.7    & 46.1    & 48.9     \\ \hline
		IRT w/ VG($A^o$)          & 38.7    & 47.5    & 50.3     \\ \hline
		IRT w/ VG($A^p$)          & 39.4    & 48.0    & 50.7     \\ \hline
		IRT w/ VG($A^o$ + $A^p$) & \textbf{39.9}    & \textbf{48.2}    & \textbf{51.0}  
		\\ \hline  
	\end{tabular}
	\label{tab.ablation_spkvg1}
\end{table}
}
\begin{table*}[t]
	\begin{center}	
		\caption{Experimental results on the Visual Genome dataset. SL denotes supervised learning. RL denotes reinforcement learning.}
		\scalebox{0.86}{	
		\begin{tabular}{c|c|c|c|c|c|c|c|c|c|c}
			
			\hline
			\multicolumn{1}{l|}{\multirow{2}{*}{Learning Strategy}}    &
			\multirow{2}{*}{Methods}  & \multicolumn{3}{c|}{PredCls}                    & \multicolumn{3}{c|}{SGClS}                    & \multicolumn{3}{c}{SGDet}                  \\ \cline{3-11} 
			\multicolumn{1}{l|}{}  
			& & R@20          & R@50          & R@100         & R@20          & R@50          & R@100         & R@20          & R@50          & R@100         \\ \hline
			\multirow{10}{*}{SL}  
			
			& VRD \cite{scenegraph:VRD_LP}
			&   -        & 27.9       & 35.0                  
			&    -       & 11.8       & 14.1      
			&   -        & 0.3        & 0.5                  \\ \cline{2-11}
			& IMP \cite{scenegraph:IMP}      
			& -          & 44.8          & 53.0          
			& -          & 21.7          & 24.4     
			& -         & 3.4          & 4.2    \\ \cline{2-11}
			& IMP+ \cite{scenegraph:IMP,scenegraph:motifs}     
			& 52.7          & 59.3          & 61.3                   
			& 31.7          & 34.6          & 35.4           
			& 14.6          & 20.7          & 24.5  \\ \cline{2-11}
			& TFR \cite{scenegraph:TFR}        
			& 40.1          & 51.9          & 58.3          
			& 19.6          & 24.3          & 26.6         
			& 3.4           & 4.8           & 6.0           \\ \cline{2-11}
			& AE \cite{scenegraph:PGAE}   
			& 47.9          & 54.1          & 55.4                  
			& 18.2          & 21.8          & 22.6                
			& 6.5           & 8.1           & 8.2   \\ \cline{2-11}
			& FREQ+OVERLAP \cite{scenegraph:motifs}        
			& 53.6          & 60.6          & 62.2                 
			& 29.3          & 32.3          & 32.9        
			& 20.1          & 26.2          & 30.1          \\ \cline{2-11}
			& Graph R-CNN \cite{scenegraph:graphrcnn}   
			& -             & 54.2          & 59.1       
			& -             & 29.6          & 31.6            
			& -             & 11.4          & 13.7            \\ \cline{2-11}
			
			& KERN \cite{scenegraph:KERN}         
			& -             & 65.8          & 67.6                 
			& -             & 36.7          & 37.4    
			& -             & 27.1          & 29.8               \\ \cline{2-11}
			& MotifNet \cite{scenegraph:motifs}        
			& 58.5          & 65.2          & 67.1            
			& 32.9          & 35.8          & 36.5     
			& 21.4          & 27.2          & 30.3                  \\
			\cline{2-11} 			
			& IRT (Ours)                                    
			& \textbf{60.3}   & \textbf{66.8}   & \textbf{68.5} 
			& \textbf{33.9}   & \textbf{36.9}   & \textbf{37.5}      
			& \textbf{22.0}   & \textbf{27.9}  & \textbf{31.1} 
			\\
			
			\cline{2-11}   
			
			& {IRT+MSK (Ours)}                                    
			& \textbf{60.4}   & \textbf{67.0}   & \textbf{68.6} 
			& \textbf{34.2}   & \textbf{37.1}   & \textbf{37.7}      
			& \textbf{22.2}   & \textbf{28.0}  & \textbf{31.2}  
			\\  
			\cline{2-11} 			
			\hline \hline
			\multirow{2}{*}{SL+RL}  
			& TreeLSTM+RLrefine \cite{scenegraph:treelstm}
			& 60.1  & 66.4  & 68.1             
			& 35.2  & 38.1  & 38.8 
			& 22.0  & 27.9  & 31.3			
			\\ \cline{2-11} 
			& CMAT+RLrefine \cite{scenegraph:cmat}                
			& 60.2  & 66.4 & 68.1			          
			& 35.9  & 39.0  & 39.8 	         
			& 22.1  & 27.9  & 31.2 			           \\ \hline 
		\end{tabular}}
		\label{tab.baseline}
	\end{center}
\end{table*}

\textbf{The roles of the Commonsense Knowledge and the Relational knowledge.}
In order to further investigate the effect of the structure knowledge, we gradually add different types of knowledge to IRT, as shown in Table~\ref{tab.ablation_spk}. Without any additional knowledge, IRT achieves presentable results, which indicate that IRT can take advantage of the contextual information to predict relationships. After adding the Relational Knowledge (RK) for IRT, the results are improved, e.g., the result increases by 2.2\% on R@20. This shows that the relational knowledge features are effectively extracted and utilized in our model. Finally, the model achieves the best, when the Commonsense Knowledge (CK) is further added. These results indicate that providing the Relational Knowledge and the Commonsense Knowledge for IRT can make up for the lack of supervision information of the OSSGG task to some extent.

\textbf{The roles of $A^o$ and $A^p$ in Visual Genome.}
Furthermore, we explore the influence of the structured knowledge from Visual Genome. As mentioned in Section~\ref{sec.spk}, two adjacency matrices ($A^o$ and $A^p$) are used to encode the structured knowledge from Visual Genome. $A^o$ is the adjacency matrix with instance categories mentioned in Section~\ref{sec.spk}. $A^p$ is the adjacency matrix with predicate categories. Table~\ref{tab.ablation_spkvg} shows that both matrices improve the effectiveness of the model, e.g., $A^o$ improves the result of IRT by $1.0\%$ on R@20, and $A^p$ further improves the result by $1.2\%$. This shows that the model requires not only the prior knowledge from instance categories ($A^o$) but also the prior knowledge from relationship predicates ($A^p$).

\footnotetext[1]{\label{foot1} We use the codes provided by Zellers~\etal~\cite{scenegraph:motifs} and train the MotifNet model on the One-Shot Visual Genome dataset. Github: https://github.com/rowanz/neural-motifs}
\subsection{Experimental Results on the Scene Graph Generation task}
In this section, we evaluate our method on the scene graph generation task to illustrate the universality of our approach. We compare our method with previous methods, including: methods adopting supervised learning~\cite{scenegraph:VRD_LP,scenegraph:IMP,scenegraph:motifs,scenegraph:TFR,scenegraph:PGAE,scenegraph:graphrcnn,scenegraph:KERN}, and methods adopting reinforcement learning~\cite{scenegraph:treelstm,scenegraph:cmat}, as shown in Table~\ref{tab.baseline}.

From Table~\ref{tab.baseline}, our IRT outperforms the state-of-the-art methods with the supervised learning strategy, such as KERN~\cite{scenegraph:KERN}, MotifNet~\cite{scenegraph:motifs}, and IMP~\cite{scenegraph:IMP}. This shows that our IRT can fully extract and utilize the contextual information among instances to assist the relationship prediction. After adding Multiple Structured Knowledge, the performances are also improved. However, the improvement is marginal. Because the task already has rich annotations that allow the model to learn stable knowledge, the additional knowledge is trivial for the scene graph generation task.

We also compare our method with the methods using reinforcement learning~\cite{scenegraph:treelstm,scenegraph:cmat} and observe the following results.
Firstly, our method achieves comparable results on the SGDet setup. The setup depends heavily on the results of object detection. Because the focus of our method and the previous methods is not the object detection task, we get similar results on the SGDet setup. 
Secondly, the methods based on the reinforcement learning strategy get better results on the SGCls because the SGCls setup focuses more on the instances classification results than the PredCls setup. \cite{scenegraph:treelstm,scenegraph:cmat} use the reinforcement learning strategy to enforce a high penalty on the misclassification of prominent instances, resulting in a better performance on the SGCls setup. 
Thirdly, on the PredCls setup, our method is better than reinforcement learning based models since the PredCls setup ignores the effect of instance categories. These results show the superiority of our method for the relationship predicate prediction. 
Moreover, compared with these methods, the training process of our method is more concise. 

\begin{figure*}[t]
	\begin{center}
		\includegraphics[width=0.75\linewidth]{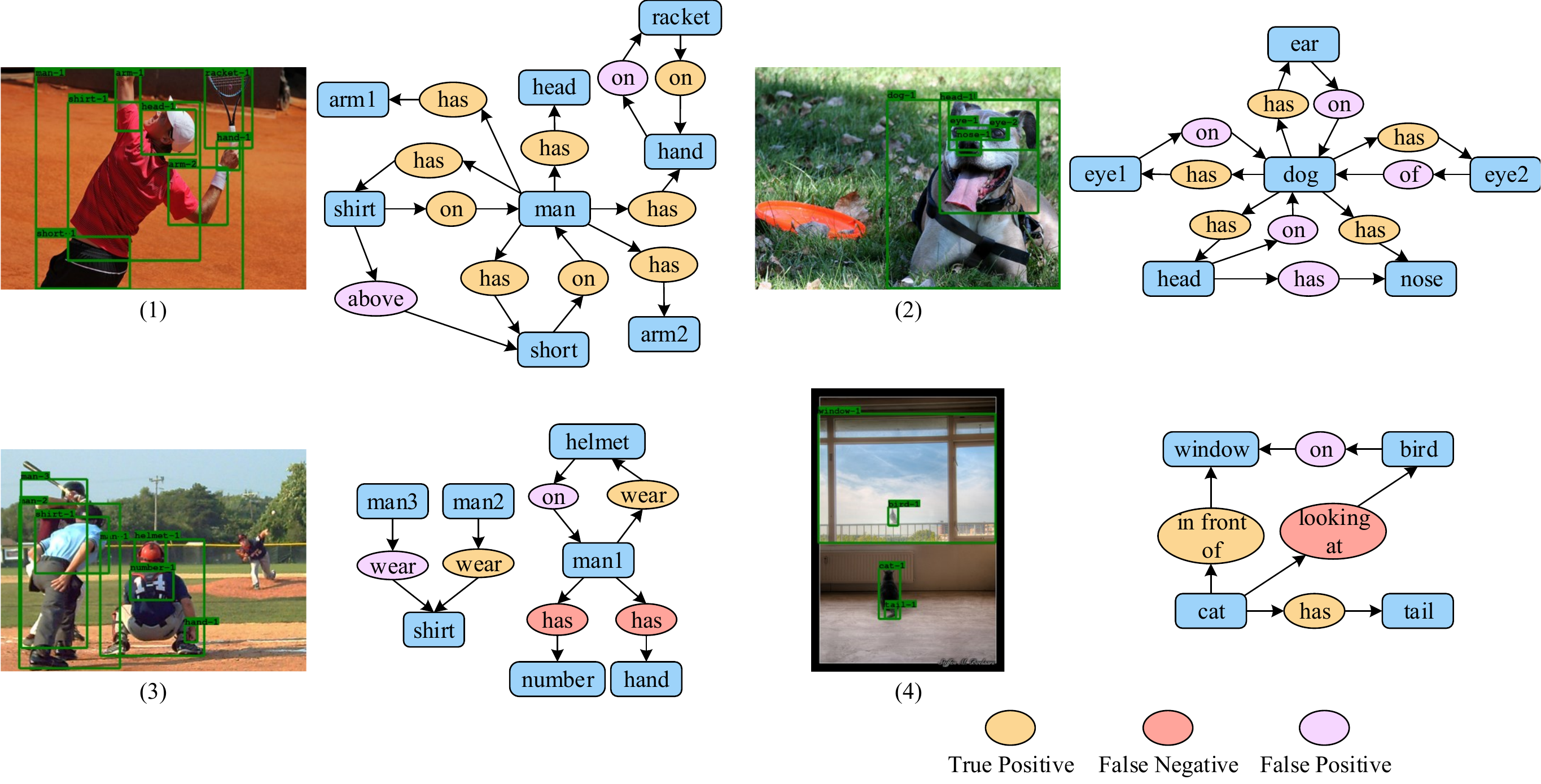}
	\end{center}
	\caption{The visualization results of Instance Relation Transformer (VGG)+MSK on the one-shot scene graph generation task. These results are generated on the PredCls setup.}
	\label{fig:quali_res}
\end{figure*}
\subsection{Qualitative Results}
In this section, we show some qualitative results in Figure~\ref{fig:quali_res}. 
With the Relational Knowledge and the Commonsense Knowledge, our IRT predicts some spatial relationships correctly, e.g., <shirt, above, shot> and <racker, on, hand>, in the first example, and some commonsense relationships, e.g., <dog, has, head> and <dog, has nose>, in the second example. We also show some notable failures in the third and fourth examples. In the third example, due to the excessive overlap between $man2$ and $man3$, the model outputs that $man3$ wears the T-shirt of $man2$. In the fourth example, the model outputs <bird, on, window> because the bounding box of the window completely contains the bird, which is a strong signal for the relationship predicate ``on''. These two failure examples are due to the bias of spatial information. Properly reducing the dependence on spatial information may help to alleviate this problem.
\section{Conclusion}
In this paper, to equip the model with the ability to learn the visual relationship from one labeled sample, we design a novel task, namely one-shot scene graph generation. Motivated by the way humans learn visual relationships, the Multiple Structured Knowledge (Relational Knowledge and Commonsense Knowledge) is introduced into the one-shot scene graph generation task. The Relational Knowledge extracted from Visual Genome represents the prior knowledge of relationships among entities in the visual space. The Commonsense Knowledge explores ``sense-making'' knowledge from ConceptNet. Besides, we propose the Instance Relation Transformer for capturing the relational context among instances. Detailed experiments validate the effectiveness of the Instance Relation Transformer and the Multiple Structured Knowledge.
\vspace{-3mm}
\begin{acks}
This work is supported by the Fundamental Research Funds for the Central Universities (Grant No. ZYGX2019J073), the National Natural Science Foundation of China (Grant No. 61772116, No. 61872064, No.61632007, No. 61602049), Sichuan Science and Technology Program (Grant No. 2019JDTD0005), The Open Project of Zhejiang Lab (Grant No. 2019KD0AB05).
\end{acks}
\bibliographystyle{ACM-Reference-Format}
\balance
\bibliography{sample-base}

\end{document}